\documentclass{article}

\usepackage{PRIMEarxiv}

\usepackage[utf8]{inputenc} 
\usepackage[T1]{fontenc}    
\usepackage{hyperref}       
\usepackage{url}            
\usepackage{booktabs}       
\usepackage{amsfonts}       
\usepackage{nicefrac}       
\usepackage{microtype}      
\usepackage{lipsum}
\usepackage{graphicx}
\graphicspath{{media/}}     
\usepackage{adjustbox}
\usepackage{amsmath} 
\usepackage{subcaption}
\usepackage{caption}

\usepackage[numbers]{natbib}
\title{NEURIPS2024 Ariel Data Challenge: Characterization of Exoplanetary Atmospheres Using a Data-Centric Approach}

\author{
  Jérémie BLANCHARD \\
  \texttt{jeremie.blanchard@crim.ca} \\
  \small Computer Research Institute of Montreal 
   \And
  Lisa CASINO \\
  \texttt{lisa.casino@crim.ca} \\
  \small Computer Research Institute of Montreal 
    \And
  Jordan GIERSCHENDORF \\
  \texttt{jordan.gierschendorf@crim.ca} \\
  \small Computer Research Institute of Montreal 
}

\begin{document}
\maketitle

\begin{abstract} 
The characterization of exoplanetary atmospheres through spectral analysis is a complex challenge. The NeurIPS 2024 Ariel Data Challenge \cite{ariel-data-challenge-2024}, in collaboration with the European Space Agency's (ESA) Ariel mission \cite{tinetti2021arielenablingplanetaryscience}, provided an opportunity to explore machine learning techniques for extracting atmospheric compositions from simulated spectral data. In this work, we focus on a data-centric business approach, prioritizing generalization over competition-specific optimization. We briefly outline multiple experimental axes, including feature extraction, signal transformation, and heteroskedastic uncertainty modeling. Our experiments demonstrate that uncertainty estimation plays a crucial role in the Gaussian Log-Likelihood (GLL) score, impacting performance by several percentage points. Despite improving the GLL score by 11\%, our results highlight the inherent limitations of tabular modeling and feature engineering for this task, as well as the constraints of a business-driven approach within a Kaggle-style competition framework. Our findings emphasize the trade-offs between model simplicity, interpretability, and generalization in astrophysical data analysis.
\end{abstract}

\section{Introduction}

The characterization of exoplanetary atmospheres is a fundamental challenge in modern astrophysics. The ability to extract precise chemical compositions from observed spectra is crucial to understanding planetary formation, evolution, and the potential habitability of distant worlds. The NeurIPS 2024 Ariel Data Challenge, organized in collaboration with the European Space Agency’s Ariel mission, presents a unique opportunity to explore data-centric methodologies \cite{NEURIPS2023_112db882, jakubik2024datacentricartificialintelligence, 9754503} for analyzing exoplanetary atmospheres. This competition challenges researchers to develop algorithms capable of extracting meaningful atmospheric compositions from simulated spectral observations, which are inherently noisy and constrained by real-world instrumental limitations.

Drawing from our experience as an applied research centre, our team implemented a data-centric approach to address this challenge. Our business-oriented methodology adheres to best practices, aiming to achieve consistent performance across both private and public datasets, even when the test set is unknown and inherently out of distribution. This approach led to the development of a robust and generalizable solution, avoiding any influence from reverse engineering of the process used to generate the data. We focus on feature engineering and data quality enhancement to improve the reliability of machine learning models.

To streamline our submission process, we developed a pipeline that enabled rapid iteration across different approaches. This paper presents our methodological framework, highlighting key insights from our feature engineering efforts and data refinement strategies. Although our solution did not rank among the top-performing models in the competition, our work offers valuable perspectives on the impact of data quality and pre-processing on model generalization. We also discuss the challenges encountered and the trade-offs between complex model architectures and well-prepared data.

Through our participation in the NeurIPS 2024 Ariel Data Challenge, our aim is to highlight the importance of data-centric approaches \cite{NEURIPS2023_112db882, jakubik2024datacentricartificialintelligence, 9754503} in machine learning for exoplanetary atmospheric characterization. Data-centric AI is an approach that emphasizes improving the quality, consistency, and labeling of data rather than focusing solely on model architecture. It prioritizes systematic data curation and iteration to enhance model performance, particularly in high-variance real-world applications. The insights gained from this competition could inform future studies on optimizing retrieval methods by prioritizing data quality over model complexity.

\section{Challenge Description and Data}

\subsection{Challenge Objectives}

One of the key challenges in exoplanetary research is the extraction of atmospheric spectra from raw observational data, a critical step in enabling the chemical characterization of distant worlds.

In this competition, synthetic data are collected by simulated instruments on board spacecraft designed for space-based observations, such as ESA telescopes. During transit spectroscopy, a small fraction of starlight filters through the atmosphere of a planet before being captured by these instruments, carrying valuable spectral signatures of its composition. However, these signals are often buried in instrument noise, with jitter noise—caused by spacecraft microvibrations—being a major limiting factor \cite{tinetti2021arielenablingplanetaryscience}. This effect, comparable in magnitude to the planetary signal itself, complicates the accurate retrieval of spectral information, particularly for small exoplanets.

Participants in this competition must face the challenge of detrending sequential 2D images of the spectral focal plane, recorded over extended observational periods, to extract the planet’s transmission spectrum (283 wavelengths) along with an uncertainty estimate. This process requires advanced noise filtering and signal reconstruction techniques to mitigate instrument-related distortions while preserving the integrity of atmospheric features. Developing robust approaches to this task is essential for ensuring that future space missions, including Ariel, can achieve their ambitious scientific goals.

The competition uses the Gaussian Log-Likelihood (GLL) score, which evaluates spectral extraction by considering both accuracy and uncertainty estimation. Unlike traditional metrics, GLL penalizes overconfident predictions, ensuring that models provide realistic confidence intervals alongside spectral reconstructions. The GLL score function is defined as:

\begin{equation} \label{eq:gll}
GLL(y, \mu_{user}, \sigma_{user}) = -\frac{1}{2} \left( \log(2\pi) + \log(\sigma_{user}^2) + \frac{(y - \mu_{user})^2}{\sigma_{user}^2} \right)
\end{equation}

where \( y \) is the ground truth pixel level spectrum, \(\mu_{user}\) is the predicted spectrum, and \(\sigma_{user}\) is the corresponding uncertainty. The GLL values are summed across all wavelengths and the entire test set to produce a final GLL value (\(L\) in equation (\ref{eq:final_score})), which is then transformed into a score using the following conversion function:

\begin{equation} \label{eq:final_score}
\text{score} = \frac{L - L_{ref}}{L_{ideal} - L_{ref}}
\end{equation}

where \( L_{ideal} \) is defined as the case where the submission perfectly matches the ground truth values, with an uncertainty of 10 parts per million (ppm), and \( L_{ref} \) is defined using the mean and variance of the training dataset as its prediction for all instances.

This metric impacts data processing by requiring:

\begin{itemize}
\item  Balancing accuracy and uncertainty to avoid penalization.
\item  Preserving spectral features while mitigating noise.
\item  Ensuring robust error estimation for reliable atmospheric characterization.
\end{itemize}

By enforcing these principles, the GLL score aligns with real-world challenges in exoplanet spectroscopy, where both signal integrity and uncertainty quantification are crucial. More information on the score is discussed in Section \ref{work_w_uncertainty}.

\subsection{Data}

\begin{figure}[h!]
    \centering
    \includegraphics[width=1\linewidth]{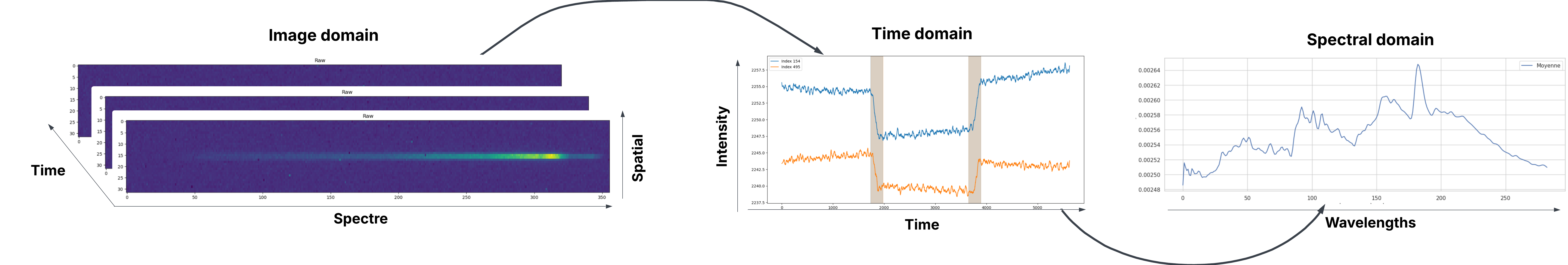}
    \caption{Illustration of the transformation applied to the raw images signal, from image domain to time domain, in order to predict the spectral domain.}.
    \label{fig:data_transformation}
\end{figure}

The Ariel Data Science Competition provided image-based datasets, which we transformed into multiple time series to better capture planetary transits, as illustrated in Figure \ref{fig:data_transformation}. This conversion was the first step in our data exploration phase and allowed us to precisely identify when an exoplanet passed in front of its star, enabling us to isolate the most relevant data segments and maximize the impact of our model. Finally, our predictions were made in the spectral domain to accurately identify the chemical composition of exoplanetary atmospheres.

The Ariel mission includes two instruments for exoplanet observation:

\begin{itemize}
\item  FGS1 (Fine Guidance System - Channel 1): A visible-light photometer (\( 0.60 \)–\( 0.80 \,\mu m\)) primarily designed for spacecraft centering and guiding, but also providing high-precision photometric measurements of the target star. This system is responsible for flux slope variations that affected some spectral measurements (more details in Section \ref{work_w_noisy_data}).
\item  AIRS-CH0 (Ariel InfraRed Spectrometer - Channel 0): An infrared spectrometer \( 1.95 \)–\( 3.90 \,\mu m\), with a resolving power $R \approx 100$, that captures detailed atmospheric spectral features, crucial for molecular identification in exoplanetary atmospheres.
\end{itemize}

The dataset consisted of 2D images from exoplanets orbiting different types of stars, each with varying physical and spectral characteristics. However, the training dataset did not contain the exact same stellar and planetary properties as the hidden test set, introducing an out-of-distribution (OOD) generalization challenge.

To mitigate this issue, we implemented a K-fold cross-validation strategy, ensuring that each fold was validated across the two distinct stars available in the dataset. To evaluate performance of models, first we analyzed the GLL score for each star and each fold. We then computed a single GLL value by macro-averaging the scores across all folds. This cross-validation GLL score (results are presented in Section \ref{results_table}) provided a global view of the model's performance across the dataset.

This approach was designed to mitigate the consequences of an out-of-distribution sample. The goal was to ensure that the computed uncertainty could generalize to new stars and was not overly specialized to specific characteristics. Indeed, the hidden test dataset contained stars with previously unseen characteristics. Therefore, it was crucial for our solution to adapt to these new characteristics.

\section{Methodological Framework}

As part of our participation in the NeurIPS 2024 Ariel Data Challenge, we approached the problem from a supervised learning perspective, formulating it as a multivariate regression task. The objective was to predict, for each temporal sequence (input), a vector of outputs containing the mean values ($\mu$) and uncertainties ($\sigma$) of the atmospheric transmission spectrum associated with an exoplanet. Each model input is a tabular representation of the processed signals FGS1 and AIRS-CH0, obtained through feature extraction from the raw data.

\begin{figure}[h!]
    \centering
    \includegraphics[width=1\linewidth]{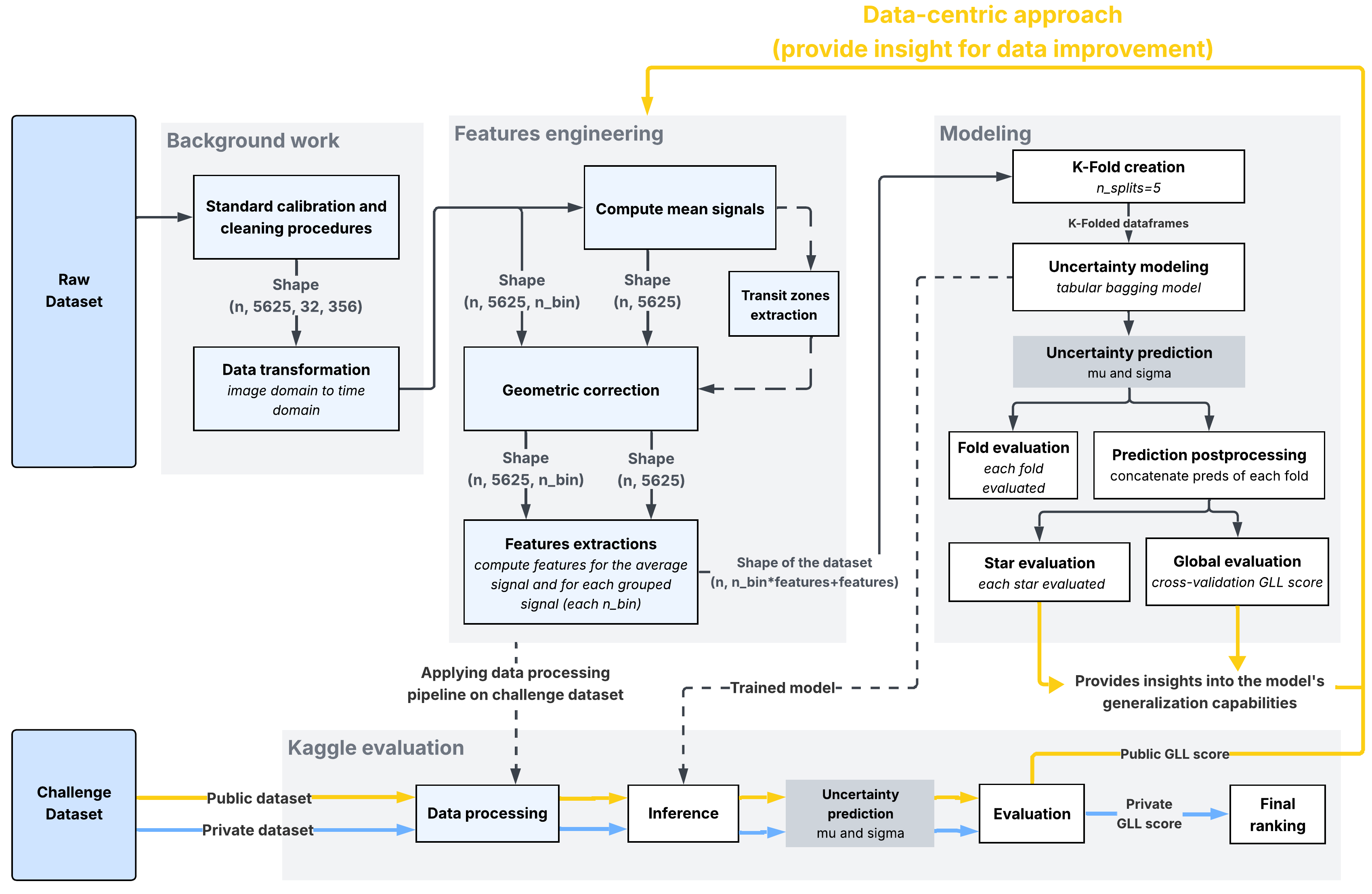}
    \caption{Overview of the proposed methodology. The workflow consists of a local data-centric pipeline including data pre-processing, feature engineering, and uncertainty-aware modeling. The trained model is then deployed on the Kaggle platform to perform inference and evaluation on the public and private challenge datasets. Insights from the public leader board and model evaluation are fed back into the pipeline (yellow arrow) to improve data processing and enhance model generalization in a data-centric approach.}
    \label{fig:process}
\end{figure}

Figure \ref{fig:process} illustrates the structure of our pipeline, which can be organized around the following main steps:

\begin{itemize}
\item \textbf{Data pre-processing (background work)}: standard calibration and cleaning procedures, followed by transformation from the image domain to the time domain. Further information in Section \ref{standard_calibration}.
\item \textbf{Feature engineering}: signal segmentation, geometric correction, and extraction of statistical and polynomial features from transit zones of both averaged and grouped (\textit{n\_bin}) signals. Further information in Section \ref{binning}.
\item \textbf{Modeling}: training and evaluation of bootstrap aggregating (bagging) prediction models with a K-fold cross-validation strategy on the extracted features. The models used are Ridge regression suited for tabular data. The modeling relies exclusively on simple feedforward architectures. Through bagging, modeling includes uncertainty quantification, and models predict both $\mu$ and $\sigma$ for each wavelength. Further information in Section \ref{work_w_uncertainty}.
\item \textbf{Postprocessing}: merging of fold predictions, transformation of the outputs to compute GLL scores for evaluation.
\end{itemize}

Our framework follows a \textbf{data-centric} paradigm, prioritizing improvements in data quality through extraction, correction, and transformation over model complexity. As shown by the yellow arrows in Figure \ref{fig:process}, feedback from the model’s performance on Kaggle’s public leader board was used to iteratively refine the upstream data processing.

This tabular feature-based approach offers good transparency and interpretability, while also being well-suited for handling uncertainty and generalization challenges associated with out-of-distribution test sets. The pipeline was also designed for reproducibility, thanks to k-fold cross-validation and strict separation between training, validation, and public/private test datasets.

\section{Experimental Results} \label{results_table}

\begin{center}
\begin{table}[h!]
    \centering
    \renewcommand{\arraystretch}{1.3}
    \begin{subtable}{\textwidth}
    \centering
    \begin{adjustbox}{width=\columnwidth,center}
    \begin{tabular}{|l||ccccccc|}
        \hline
        \textbf{Iteration} & 
        \textbf{n°1 (Baseline)} & 
        \textbf{n°2} & 
        \textbf{n°3} & 
        \textbf{n°4} & 
        \textbf{n°5} & 
        \textbf{n°6} &
        \textbf{n°7 (Final)} \\
        
        \hline
        \textbf{Cross-validation GLL score} & 
        35.23\% & 54.94\% & 36.67\% & 54.94\% & 59.94\% & 55.85\% & \textbf{66.38\%} \\
        
        \hline
        \textbf{Mean uncertainty ($\sigma_{user}$)} & 
        23.5e-05 & 7.6e-05 & 21.7e-05 & 7.6e-05 & 5.7e-05 & 2.7e-05 & 3.7e-05 \\
        
        \hline
        \textbf{Public GLL score} & 
        38.96\% & 39.92\% & 39.05\% & 45.38\% & 45.34\% & 46.69\% & \textbf{49.41\%} \\
        
        \hline
        \textbf{Private GLL score} & 
        38.44\% & 32.73\% & 38.80\% & 41.83\% & 39.69\% & 45.12\% & \textbf{49.93\%} \\
        \hline
    \end{tabular}
    \end{adjustbox}
    \caption{Comparison of different modeling iterations.}
    \label{tab:results1}
    \end{subtable}
    
    \vspace{1em} 
    
    \begin{subtable}{\textwidth}
    \centering
    \begin{adjustbox}{width=\textwidth,center}
    \begin{tabular}{|l||ccccccc|}
        \hline
        \textbf{Contributions} & 
        \textbf{n°1 (Baseline)} & 
        \textbf{n°2} & 
        \textbf{n°3} & 
        \textbf{n°4} & 
        \textbf{n°5} & 
        \textbf{n°6} &
        \textbf{n°7 (Final)} \\
        
        \hline
        \textbf{Calibration} & 
        \checkmark & \checkmark & \checkmark & \checkmark & \checkmark & \checkmark & \checkmark \\
        
        \textbf{Spectral bands (bins)} & 
        & \checkmark (5) & \checkmark (8) & \checkmark (8) & \checkmark (8) & \checkmark (8) & \checkmark (10) \\
        
        \textbf{Geometric signal correction} & 
        & & & & \checkmark & & \\
        
        \textbf{Target transformation} & 
        & \checkmark & &  \checkmark & \checkmark & \checkmark & \checkmark \\

        \textbf{Feature engineering} & 
        ~ (10) & \checkmark (158) & \checkmark (42) & \checkmark (288) & \checkmark (288) & \checkmark (288) & \checkmark (288) \\

        \textbf{Ensemble-based uncertainty quantification} & 
        & & & &  & \checkmark & \checkmark \\
        
        \hline
    \end{tabular}
    \end{adjustbox}
    \caption{Summary of contributions in different modeling iterations.}
    \label{tab:results2}
    \end{subtable}
    
    \caption{Performance comparison (top) and contribution summary (bottom) of the different iterations submitted to the NeurIPS 2024 Ariel Challenge. In the table (a), three GLL scores are shown: \textit{Cross-validation GLL} (based on the cross-validation strategy), \textit{Public GLL score} (public Kaggle leader board) and \textit{Private GLL score} (private leader board after the competition end). The \textit{Mean uncertainty (sigma)} metric refers to the average predicted uncertainty across all models. The table (b) summarizes the key contributions of each iteration: calibration, the number of bins (the raw signals were divided into multiple bins for feature extraction), geometric signal correction, target transformation (to stabilize predictions, applied in most iterations), feature engineering (the number of features created) and heteroskedastic uncertainty (introduced in the last two iterations to better capture prediction uncertainty).}
    \label{tab:results_full}
\end{table}
\end{center}

The results of our experiments presented in the table \ref{tab:results_full}  highlight several important findings regarding the different modeling iterations. The simplest models with the fewest contributions, iterations 1 and 3, were the only ones to achieve better public and private GLL scores on Kaggle than the cross-validation scores. This indicates that their simpler design may have contributed to better generalization to the test data, reducing the risk of overfitting on the validation set. As it turned out, we faced considerable difficulty adapting to out-of-distribution data, particularly with iteration 2. This model showed a significant decline in both public GLL score (40\%) and private GLL score (33\%) compared to the validation one (55\%). From iteration 1 to iteration 2, we increased the number of features from only 10 to 158 by incorporating more detailed spectral and temporal information from the signal. Moreover, the target transformation was introduced to enhance the model’s ability to capture complex patterns. However, the added complexity did not result in improved performance, leading us to revert to a simpler iteration in iteration 3. This allowed us to reassess and validate the contributions more carefully, addressing each one individually to identify the most impactful factors for model performance. Between iterations 3 and 4, a focus on data-centric optimization through feature engineering helped achieve a public GLL score over 45\% on Kaggle. Once again, this improvement came at the cost of overfitting, as demonstrated by the performance gap between the cross-validation GLL and the Kaggle leader board GLL scores. In iteration 5, the application of geometric signal correction techniques did not lead to a significant improvement. Finally, the introduction of heteroskedastic sigma estimation with uncertainty quantification in iterations 6 and 7 brought a substantial improvement in model performance, particularly in the private GLL scores. The final model integrated several important improvements, including the optimization of the tabular model, the addition of two bins to refine feature precision, and the fine-tuning of sigma. These combined adjustments led to a substantial performance boost, achieving a private GLL score of 50\%, which was the highest score among all submissions.

\section{Contributions}

This section presents our main contributions on this competition by presenting how we handled the data and and how we manage the uncertainty.

\subsection{Working with Noisy Data} \label{work_w_noisy_data}

\subsubsection{Standard Calibration and Cleaning Procedures} \label{standard_calibration}

This section is prior work done by the competition Authors and was used globally by all the competitors.

Raw data undergoes multiple pre-processing steps to correct for instrumental effects and improve signal quality before further analysis. The first step involves reverting the Analog-to-Digital Conversion (ADC) applied by the detector, using gain and offset\footnote{Offset calibration in analog-to-digital conversion (ADC) is the process of compensating for systematic voltage offsets that cause inaccuracies in the digital output. These offsets arise from imperfections in the ADC circuitry, such as input bias currents, mismatched components, or thermal variations.} calibration values to retrieve a physically meaningful signal.

Next, a masking operation is applied to filter out dead and hot pixels, ensuring that non-responsive pixels do not interfere with subsequent calculations. These masked pixels remain uncorrected, as further interpolation methods are left for later processing. The pixel response non-linearity is then corrected by applying an inverse polynomial transformation derived from calibration coefficients, compensating for capacitive leakage effects in the detector electronics. Following this, dark current subtraction is performed to remove the background signal accumulated in each pixel during exposure, using precomputed dark frames. The data is further refined through Correlated Double Sampling (CDS) \cite{donati2000photodetectors}, where the detector is read twice—once at the start and once at the end of the exposure—to remove readout noise by taking their difference. Optionally, time-binning can be applied to reduce data volume while preserving relevant information. Finally, a flat-field correction is performed to account for pixel-to-pixel variations in detector sensitivity, normalizing the response across the image and ensuring uniformity in spectral measurements. These corrections provide a cleaner dataset, facilitating accurate characterization of exoplanetary atmospheres. \footnote{Notebook from Kaggle: https://www.kaggle.com/code/gordonyip/update-calibrating-and-binning-astronomical-data. This notebook is prepared by Angèle Syty and Virginie Batista (IAP), with support from Andrea Bocchieri, Orphée Faucoz (CNES), Lorenzo V. Mugnai (Cardiff University \& UCL), Tara Tahseen (UCL), Gordon Yip (UCL).}

\subsubsection{Preserving Spectral Features Through Binning} \label{binning}

\begin{figure}[h!]
    \centering
    \begin{subfigure}[b]{0.45\textwidth}
        \centering
        \includegraphics[width=\textwidth]{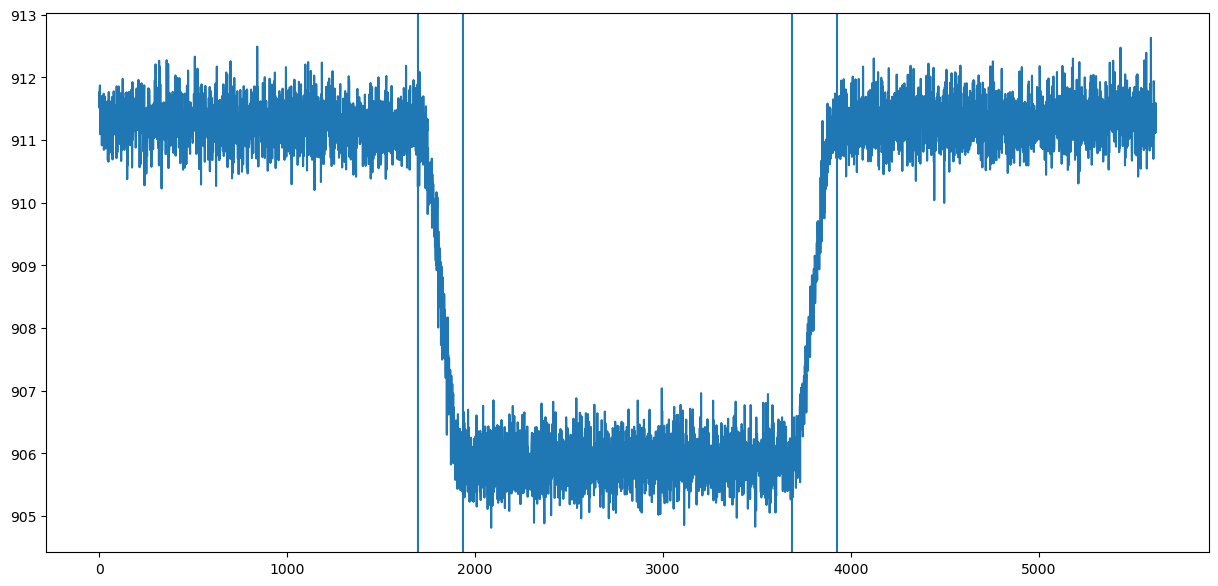}
        \caption{Mean AIRS signal}
        \label{fig:airs_mean}
    \end{subfigure}
    \hfill
    \begin{subfigure}[b]{0.45\textwidth}
        \centering
        \includegraphics[width=\textwidth]{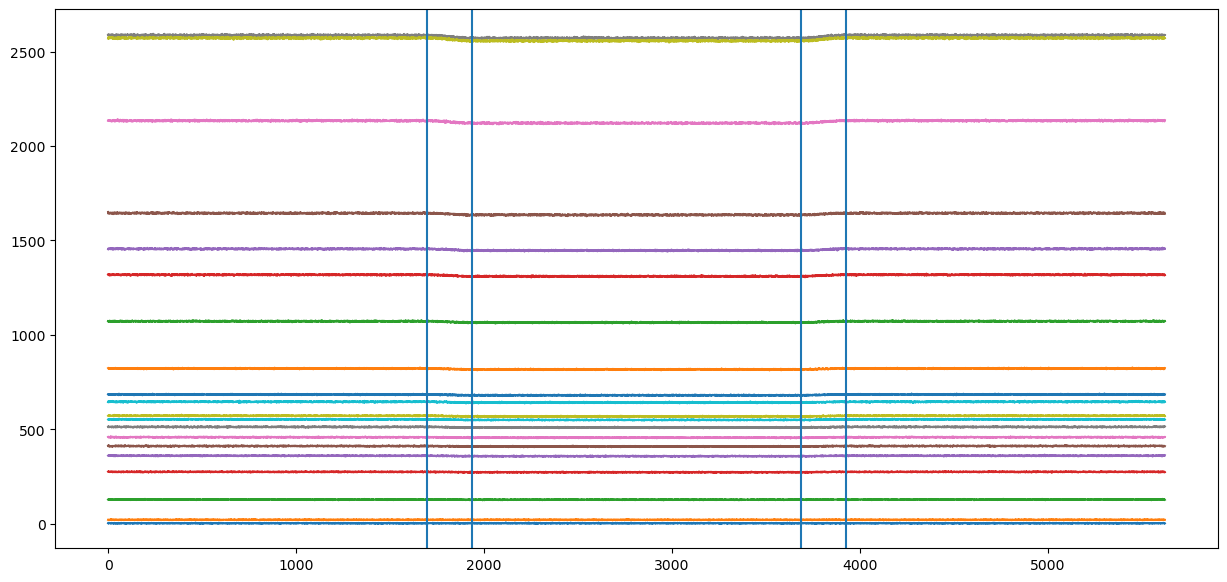}
        \caption{Binning AIRS signal}
        \label{fig:airs_bins}
    \end{subfigure}
    \caption{Mean Representation and Binning of AIRS signal}
    \label{fig:airs}
\end{figure}

The original AIRS data is structured as a 3D dataset, with dimensions representing time, spatial coordinates, and spectral information. The spectral axis consists of 356 values corresponding to different wavelengths. A straightforward yet aggressive transformation involves averaging across the spectral dimension, reducing the data to a 2D representation (time and spatial dimensions) as illustrated in Figure \ref{fig:airs_mean}. While this simplification facilitates analysis, it also results in a significant loss of spectral detail, which is crucial for characterizing exoplanetary atmospheres.

To preserve key spectral features while reducing data complexity, we implemented a binning iteration along the spectral axis instead of performing a full averaging, shown in Figure \ref{fig:airs_bins}. This method divides the 356 spectral values into a predefined number of bins $n$, typically 5 to 30, depending on the experiment. Each bin groups a set of contiguous wavelengths, and the values within each bin are averaged. The exact number of wavelengths grouped in each bin is calculated as the floor division of the total spectral values (356) by the number of bins $n$. If there is a remainder, the final bin includes the remaining wavelengths. An example is shown in Figure \ref{fig:airs_bins}. This method retains critical information related to chemical compositions, temperature profiles, and potential biosignatures while reducing data complexity. By averaging the spectral data within each bin, we also reduce noise, particularly the "jitter noise" caused by spacecraft vibrations. This iteration strikes a balance between spectral detail retention and data simplification, making it especially relevant for large-scale datasets expected from missions like ARIEL. The resulting binned data retains essential spectral characteristics while ensuring efficient processing.

\subsubsection{Geometrically Correcting AIRS Spectral Signals} \label{noisy_data}

\begin{figure}[h!]
    \centering
    \includegraphics[width=0.8\linewidth]{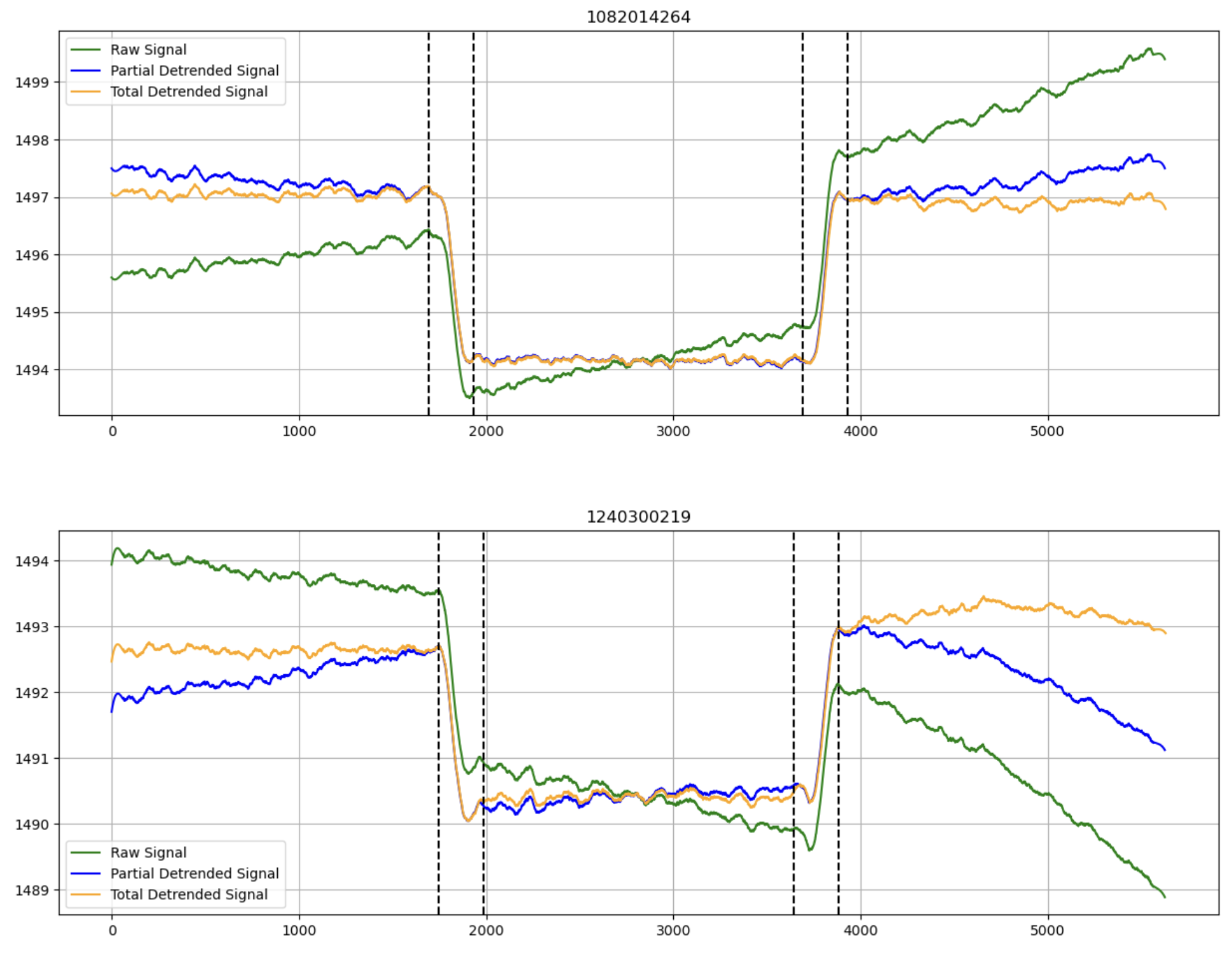}
    \caption{Illustration of the signal correction process. The green line represents the raw signal obtained from the AIRS instrument, showing the initial data with potential noise and drifts. The blue line indicates the local detrended signal, where initial corrections have been applied to mitigate trends and noise. The orange line shows the global corrected signal after applying slope and trend adjustments across all segments.}
    \label{fig:segment}
\end{figure}

AIRS spectral signals frequently exhibit consistent temporal slopes that can affect the interpretation of exoplanetary transit events. To address this, we applied a geometric correction by first identifying breakpoints corresponding to the ingress and egress zones of the transit (refer to the vertical dashed lines in Figure \ref{fig:segment}). These breakpoints were used to segment the signal into three distinct flat zones: pre-transit (left), in-transit (middle), and post-transit (right). This iteration allowed us to isolate portions of the signal where the variations were more stable.

Within each flat zone, we estimated and corrected slopes by calculating and removing the linear trend from each segment. This step reduced systematic biases and drifts that could otherwise distort the observed spectral variations. A global adjustment was then applied using the middle flat zone as a reference, aligning all segments to a common baseline. By applying this correction, the processed signals provide a more consistent representation of spectral features, improving the reliability of atmospheric characterizations for exoplanets observed with ARIEL.

However, despite the improvement in signal consistency, we observed a slight degradation in performance on the test set. One possible hypothesis for this phenomenon is that the geometric correction, while effective at removing trends and stabilizing the signal in the training data, may inadvertently remove subtle but important variations in the test data, which could be linked to real physical processes in different star systems. This could lead to a loss of crucial information that the model relies on for accurate predictions, particularly for out-of-distribution data. Moreover, the correction could introduce a form of overfitting to the training data specific characteristics, reducing the model ability to generalize to unseen conditions.

\subsubsection{Target Scaling}

In our methodology, we apply a standard scaler to the target variables to standardize their distribution. This technique involves transforming the targets by subtracting the mean and dividing by the standard deviation, resulting in targets with a mean of zero and a unit variance. This step is particularly useful when working with models sensitive to the scale of the data, such as linear regression or neural networks, as it helps to improve convergence and stability during training. Standard scaling of targets also facilitates better comparison between models, especially when the targets vary across different ranges or units.

\subsubsection{Features Extraction from Spectral Signals}

Feature engineering was applied to the spectral signals to extract meaningful information for exoplanetary transit analysis. The first step involved segmenting the signals into five distinct regions: left, middle, and right flat zones, along with ingress and egress transition zones, described in Section \ref{noisy_data}. This segmentation allowed for a more detailed examination of the different phases of the transit event, particularly the transition zones where light intensity variations encode valuable atmospheric properties. Statistical features such as the minimum, maximum, mean, and standard deviation were extracted from these transition regions to characterize the dynamics of the exoplanet’s passage across the stellar disk. Additionally, polynomial functions of degrees 3 and 4 were fitted to the transition zones, with their coefficients included as features to capture subtle variations in the signal. This iteration helped reduce the impact of observational noise, including spacecraft "jitter noise," which can obscure faint atmospheric signatures. Features were computed not only on the mean signal but also on binned signals, ensuring that both broad trends and fine spectral details were preserved in the analysis. Furthermore, the same features were extracted from the Fine Guidance Sensor (FGS) signals to complement the spectral data and provide additional validation for the observed patterns. By incorporating statistical, polynomial, and multi-scale spectral features, this feature engineering iteration maximized the extraction of scientifically relevant information from the transit signals.

\subsubsection{Working with Uncertainty} \label{work_w_uncertainty}

\begin{figure}[h!]
    \centering
    \includegraphics[width=0.55\linewidth]{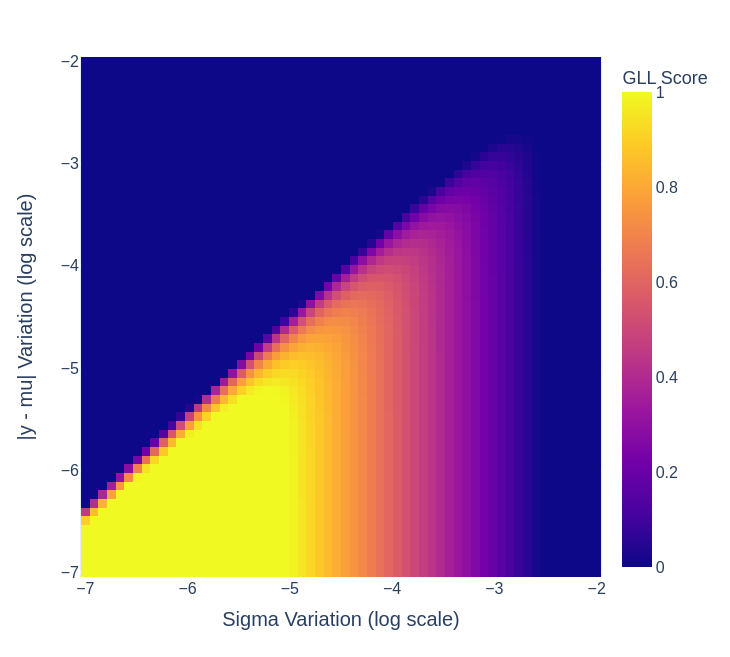}
    \caption{Variation of the Gaussian Log-Likelihood (GLL) score defined in the equation (\ref{eq:final_score}) with respect to the \(\sigma_{user}\) uncertainty and prediction (\(\mu_{user}\)) values. The heatmap illustrates how different combinations of \(\mu_{user}\) and \(\sigma_{user}\) impact the final score. When the uncertainty (\(\sigma_{user}\)) is smaller than the prediction error, the score drastically decreases, as seen in the region above the diagonal. Conversely, when uncertainty is larger than the prediction error, the model is more tolerant, leading to a higher score. The region below the diagonal shows this behavior.}
    \label{fig:score}
\end{figure}

The precision of the predicted uncertainty significantly impacts the GLL score. When uncertainty is high, the quadratic error term in (\ref{eq:gll}) has a reduced impact, reducing the penalty for prediction errors and allowing the model to be more tolerant of differences between the predicted \(\mu_{user}\) and observed spectra \(y\). This can be beneficial when the data are noisy or when the model predictions are less accurate. However, the log-variance term in (\ref{eq:gll}) becomes more dominant, resulting to a lower overall score. In contrast, when uncertainty is low the GLL score becomes highly sensitive to prediction errors, leading to higher scores when predictions closely match the actual data. On the other hand, inaccurate predictions are penalized more severely in this case. The variation of \(\mu_{user}\) and \(\sigma_{user}\) are illustrated in Figure \ref{fig:score}.

We implemented a custom Bootstrap Aggregating (Bagging) approach \cite{breiman96} to quantify the prediction uncertainty at each wavelength across different exoplanets. This ensemble technique enhances prediction accuracy by aggregating outputs from multiple models trained on different subsets of the data, thus promoting model diversity. Specifically, we trained 50 individual models, each using between 60\% and 80\% of the training data, sampled with replacement. This strategy helps mitigate overfitting and increases model robustness, especially when generalizing to out-of-distribution data such as unseen stars in the public and private test sets. 

Any model implemented with the base class applicable to all estimators can be used. In all our experiments presented in Table \ref{tab:results_full}, we used a Ridge Kernel. The final iteration 7 in \ref{tab:results1} uses the following parameters:

\begin{itemize}
    \item $\alpha$ : 0.0001
    \item $\gamma$ : 0.0001
    \item Kernel space : Polynomial
    \item Polynomial degree: 3
    \item Number of bagged models: 50
    \item Percentage of sampling: 80\%
\end{itemize}

\begin{figure}[h!]
    \centering
    \begin{subfigure}[b]{0.45\textwidth}
        \centering
        \includegraphics[width=\textwidth]{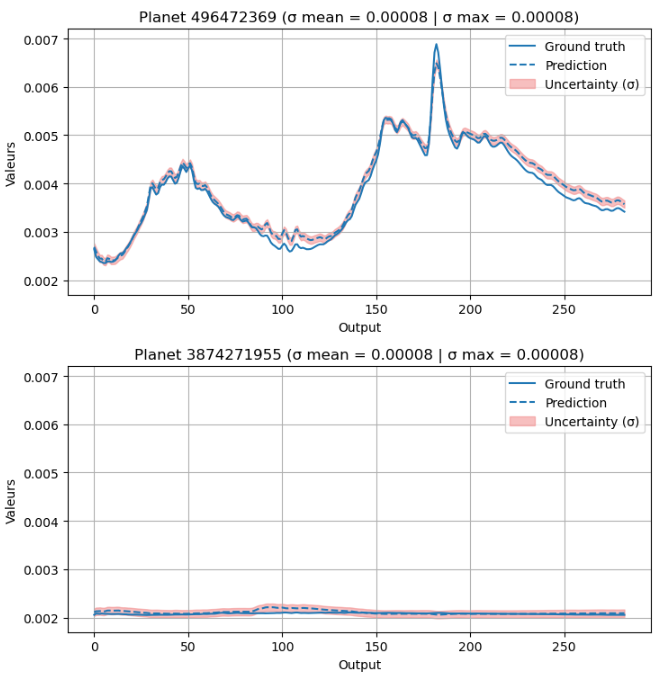}
        \caption{Predicted and actual wavelengths in approach 5, where uncertainty is fixed.}
        \label{fig:sigma_fixe}
    \end{subfigure}
    \hfill
    \begin{subfigure}[b]{0.45\textwidth}
        \centering
        \includegraphics[width=\textwidth]{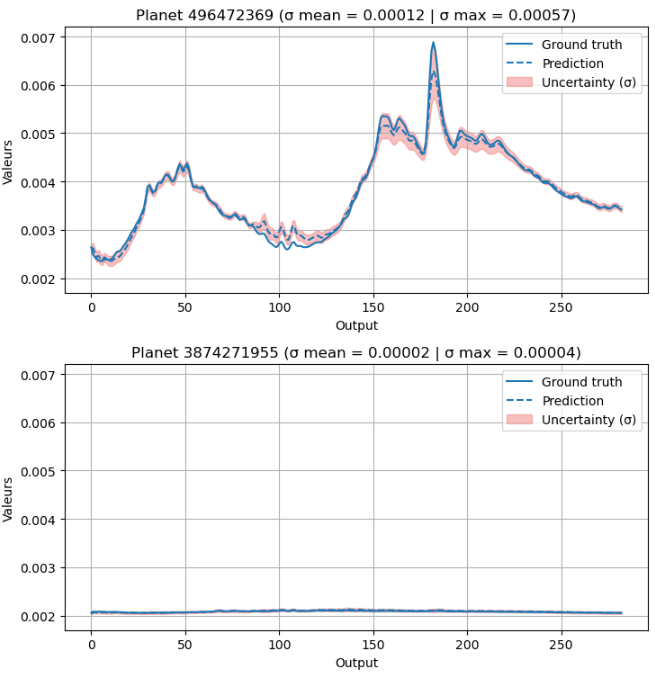}
        \caption{Predicted and actual wavelengths in approach 7, where uncertainty is heteroskedastic.}
        \label{fig:sigma_dynamic}
    \end{subfigure}
    \caption{Comparison of ground truth (\(y\)) and predictions (\(\mu_{user}\), \(\sigma_{user}\)) with fixed and heteroskedastic (i.e., data-dependent) uncertainty approaches. The fixed \(\sigma_{user}\) in the approach \ref{fig:sigma_fixe} can lead to overly large uncertainties (bottom plot) or overly small uncertainties (top plot), which penalizes the GLL score. The approach \ref{fig:sigma_dynamic} adjusts to the model predictions, resulting in better calibration of uncertainty. Higher uncertainties reflect less agreement among the models, while smaller uncertainties suggest greater consistency across model predictions.}
    \label{fig:results_two_planets}
\end{figure}

To estimate uncertainty, we compute the standard deviation of the predictions across the 50 models. A higher standard deviation indicates greater disagreement among models, and thus higher predicted uncertainty. This provides an interpretable and data-driven measure of confidence, as illustrated in Figure~\ref{fig:results_two_planets}.

\section{Teaching and Limitations}

While our data-centric approach is relevant for industrial applications, it encounters several limitations that are outlined below.

The hidden out-of-distribution test set significantly increased the challenge, as the exoplanets and stars it contained had different characteristics from those in the training data. While data augmentation using physical models could have been a potential solution, we decided not to explore this approach during the competition to avoid introducing data leakage.

In the presence of systematic out-of-distribution test samples, one of the main challenges of the competition was optimizing uncertainty quantification. In an operational system, handling out-of-distribution samples should be central to the modeling process. Addressing this challenge is fundamental to maintaining a reliable and adaptable system in real-world environments. Even if our final approach, with a cross-validation GLL score of 66\%, had achieved perfect uncertainty estimation (i.e., uncertainty matching the error), the cross-validation GLL score would have only reached 80\%. This highlights a remaining gap in accurately predicting the wavelengths, suggesting that further improvements are needed in the modeling process.

In theory, linear models generally exhibit better extrapolation capabilities when data distribution shifts occur than tree-based models \cite{bengio2010decision}. Using this principle, we decided to prioritize linear models over tree-based models to deal with systematic out-of-distribution test samples.

Additionally, experiments presented in Table \ref{tab:results_full} showed that an excessive number of features extracted through feature engineering led to overfitting and a decrease in the GLL Kaggle's score. Similarly, curve correction introduced challenges, as a lack of domain knowledge made it difficult to standardize the data while preserving the distinguishing features between different exoplanets.

Our team, primarily composed of data scientists and computer vision specialists, faced limitations due to the lack of direct access to a domain expert. This constraint made certain modeling and data interpretation steps more complex, particularly when defining physically meaningful transformations and evaluating the relevance of extracted features. Closer collaboration with astrophysicists could have provided valuable insights to refine the methodology and improve the robustness of the approach.

\section*{Acknowledgements}

We would like to express our sincere gratitude to BEAULIEU Mario - Computer Vision specialist at CRIM, SALHI Salma  and DOSHI Dhvani for their support and insightful discussions throughout this project.

We also extend our appreciation to the organizers of the NeurIPS 2024 Ariel Data Challenge and the European Space Agency (ESA) Ariel mission team for providing the dataset and fostering research in exoplanetary characterization. Ariel Data Challenge 2024 Organising Team:

\begin{itemize}
    \item Dr. Kai Hou Yip (UCL)
    \item Ms. Rebecca L. Coates (UCL)
    \item Dr. Lorenzo V. Mugnai (Cardiff University \& UCL)
    \item Dr. Andrea Bocchieri (Sapienza Università di Roma)
    \item Dr. Andreas Papageorgiou (Cardiff University)
    \item Mr. Orphée Faucoz (CNES)
    \item Ms. Tara Tahseen (UCL)
    \item Dr. Virginie Batista (IAP)
    \item Ms. Angèle Syty (IAP)
    \item Mr. Arun Nambiyath Govindan (UCL)
    \item Dr. Ingo P. Waldmann (UCL)
\end{itemize}


We would like to the thank Ministry of Economy, Innovation and Energy (MEIE) of the Government of Quebec for the continued support.

\medskip

\bibliography{references}

\end{document}